\documentclass[copyright,creativecommons]{eptcs}
\providecommand{\event}{FMAS 2022} 

\usepackage{iftex}
\usepackage{booktabs}

\ifpdf
  \usepackage{underscore}         
  \usepackage[T1]{fontenc}        
\else
  \usepackage{breakurl}           
\fi

\title{Towards Planning and Adaptation of Assistive-care Robot Tasks 
}
\title{Towards Adaptive Planning of Assistive-care Robot Tasks 
}

\author{
Jordan Hamilton
\institute{Department of Computer Science \\ University of York \\ York, United Kingdom}
\email{jordan.hamilton@york.ac.uk}
\and
Ioannis Stefanakos
\institute{Department of Computer Science \\ University of York \\ York, United Kingdom}
\email{ioannis.stefanakos@york.ac.uk}
\and
Radu Calinescu
\institute{Department of Computer Science \\ University of York \\ York, United Kingdom}
\email{radu.calinescu@york.ac.uk}
\and
Javier C\'amara
\institute{Department of Computer Science \\ University of M\'alaga \\ M\'alaga, Spain}
\email{jcamara@uma.es}
}
\def\titlerunning{Towards Adaptive Planning of Assistive-care Robot Tasks}
\def\authorrunning{Jordan Hamilton, Ioannis Stefanakos, \ \\Radu Calinescu and Javier C\'amara}

\usepackage[english]{babel}

\usepackage{graphicx}
\usepackage{soul}
\usepackage{xcolor,colortbl}
\DeclareRobustCommand{\revised}[1]{{\sethlcolor{white}\hl{#1}}}
\usepackage{enumitem}
\usepackage{amsmath,amsfonts,amssymb,amsthm,bm}
\usepackage{subfig}
\usepackage{wrapfig}
\usepackage{multirow}

\newlength{\tmpShadow}
\newcommand{\ourShadow}[2]{%
    \settowidth{\tmpShadow}{#1}
    \addtolength{\tmpShadow}{.1em}
    \raisebox{-0.25ex}{\textcolor{gray!70}{#1}}%
    \kern-\tmpShadow%
    \textcolor{#2}{#1}%
}

\usepackage{algorithm}
\usepackage{algpseudocode}
\usepackage{etoolbox}\AtBeginEnvironment{algorithmic}{\scriptsize} 

\usepackage{xparse}
\makeatletter
\NewDocumentCommand{\LeftComment}{s m}{%
  \Statex \IfBooleanF{#1}{\hspace*{\ALG@thistlm}}\(\triangleright\) #2}
\makeatother

\definecolor{lightskyblue}{rgb}{0.53, 0.81, 0.98}
\definecolor{atomictangerine}{rgb}{1.0, 0.6, 0.4}
\definecolor{darkpastelgreen}{rgb}{0.01, 0.75, 0.24}

\usepackage{tikz}
\newcommand*\circledEND[1]{\tikz[baseline=(char.base)]{
            \node[shape=circle,fill=lightskyblue,inner sep=2pt,scale=0.85,minimum size=1.5em,inner sep=1] (char) {\textcolor{black}{#1}};}}
            
\newcommand*\circledSTART[1]{\tikz[baseline=(char.base)]{
            \node[shape=circle,fill=yellow,inner sep=2pt,scale=0.85,minimum size=1.5em,inner sep=1] (char) {\textcolor{black}{#1}};}}
            
\newcommand*\circledHuman[1]{\tikz[baseline=(char.base)]{
            \node[shape=circle,fill=atomictangerine,inner sep=2pt,scale=0.85,minimum size=1.5em,inner sep=1] (char) {\textcolor{black}{#1}};}}
            
\newcommand*\circledAgent[1]{\tikz[baseline=(char.base)]{
            \node[shape=circle,fill=darkpastelgreen,inner sep=2pt,scale=0.85,minimum size=1.5em,inner sep=1] (char) {\textcolor{black}{#1}};}}
            
\newcommand*\circledConflict[1]{\tikz[baseline=(char.base)]{
            \node[shape=circle,fill,inner sep=2pt,scale=0.85,minimum size=1.5em,inner sep=1] (char) {\textcolor{white}{#1}};}}
            
\makeatletter
\renewcommand{\paragraph}{%
  \@startsection{paragraph}{4}%
  {\z@}{1ex \@plus 1ex \@minus .2ex}{-1em}%
  {\normalfont\normalsize\bfseries}%
}
\makeatother

\begin{document}
\maketitle

\begin{abstract}
This `research preview' paper introduces an adaptive path planning framework for robotic mission execution in assistive-care applications. The framework provides a graph-based environment modelling approach, with dynamic path finding performed using Dijkstra's algorithm. 
A predictive module that uses probabilistic model checking is applied to estimate the human's movement through the environment, allowing run-time re-planning of the robot's path. We illustrate the use of the framework for a simulated assistive-care case study in which a mobile robot \revised{navigates through the environment and} monitors an end user with mild physical or cognitive impairments.
\end{abstract}


\section{Introduction}

Whilst assistive robots \cite{FeilSeifer} have been embedded into social and health care environments \cite{Bedaf2017,Beuscher2017,Hebesberger2017}, they have largely been limited to simple applications, such as support for social and physical activities and hall monitoring, \revised{but often without considering potential interactions with humans}. 
To expand the range of these applications, 
the human user and the robot need to interact in order to perform tasks together~\cite{calinescu2019socio}. As such, this interaction, which 
is still underexplored 
in the social care domain, should be prioritised, 
with an emphasis on the safety of the human~\cite{calinescu2013emerging,Gleirscher2022}.

To enable the development of applications that support such interaction and to 
ensure its safety, we propose an adaptive mission and path finding framework for an autonomous robot operating in a home-care environment. The framework models the environment as a graph, with nodes representing key locations within the environment where the robot can perform local tasks. Missions are modelled as a repertoire of locations within the environment where a task requires completion. 

The main contributions of our `research preview' paper are: \textbf{(i)}~a generalised approach for modelling environments as graphs with edges represented as levels of risk, \textbf{(ii)}~a modified Dijkstra's algorithm for performing path finding in uncertain environments with a cost function to reduce risk, \textbf{(iii)}~simple human predictive behaviour model that forecasts human intention allowing for adaptive path finding using heat maps to artificially increase the risk associated with specific edges in the graph, \textbf{(iv)}~a framework that combines modelling methods, adaptive path finding techniques and run-time probabilistic model generation for safety verification into an end-to-end solution for autonomous robotic mission planning, \textbf{(v)}~finally, a simulation-based case study that shows the effectiveness of the framework. 

\section{Framework with Illustrative Example}



\begin{wrapfigure}[15]{r}{0.45\textwidth}
    \centering
    \vspace{-10pt}
    \includegraphics[width=0.43\textwidth]{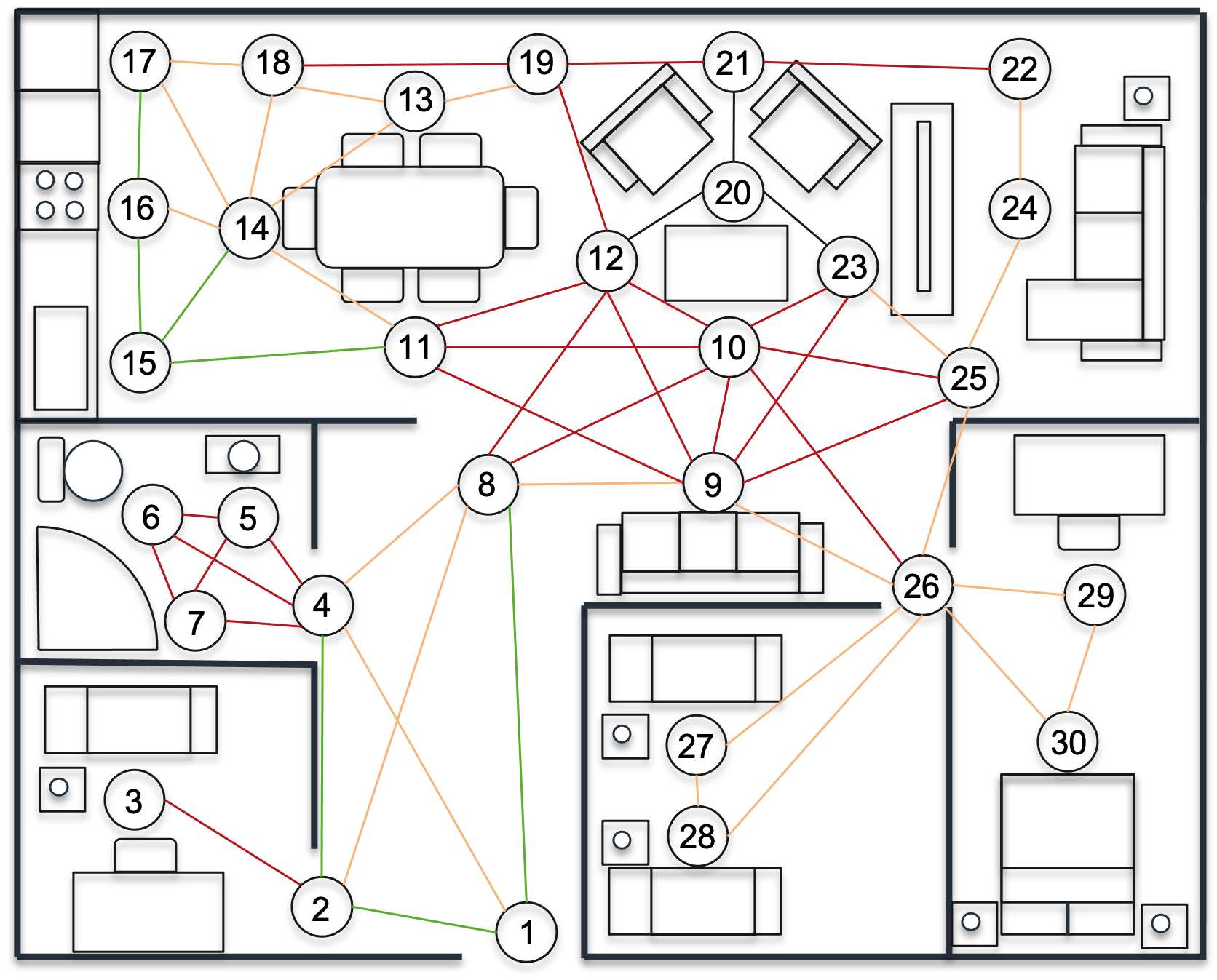}
    \caption{Multi-room environment layout with nodes and risk severity color-coded edges}
    \label{fig:MotivatingEnvironmentLayout}
\end{wrapfigure}

\paragraph{Motivating application.} \label{subsection: Motivating application}
We will illustrate the operation of our framework for an assistive-care robotic application in which a mobile robot operates in an uncertain environment, and in close proximity to an untrained user. This environment (Figure~\ref{fig:MotivatingEnvironmentLayout}) consists of five individual rooms located on the same floor with a large open plan kitchen and living area. A set of 30~nodes (i.e., locations) within this space correspond to key areas of the environment where the robot can perform specific tasks or that the robot may reach while navigating through the environment. 

The robot navigates the environment to reach a specific node for which a task is performed at this node; although we do not consider the local activity being performed at this node but rather the navigation and ordering of the tasks in this framework. Edges connect neighbouring nodes based on line-of-sight, and these edges enable movement of the robot through the environment using a path planning algorithm discussed later on. Each edge is assigned a risk metric, with risk measured as the probability of failing to traverse the edge successfully. In this example, high risk edges (depicted using red and black lines in Figure~\ref{fig:MotivatingEnvironmentLayout}) occur in cluttered spaces areas where foot traffic is highly likely. Similarly, low risk edges (depicted as green lines) are associated with uncluttered areas unlikely to be used as pathways by the user. 
%

In this paper, we consider a scenario in which the robot is responsible for monitoring the health of a frail user by periodically patrolling the environment to check his or her status. 
\revised{To perform this type of mission, the robot will need to move through the environment checking certain locations. An example of this type of mission could be to visit locations $3,\ 4,\ 9,\ 10,\ 28$ and $29$. Since the mission is time critical, the order in which operations are to be carried out is not predefined and should be determined at run-time based on the robot's starting location. 
}

\begin{figure}[!t]
    \centering
    \includegraphics[width=0.8\textwidth]{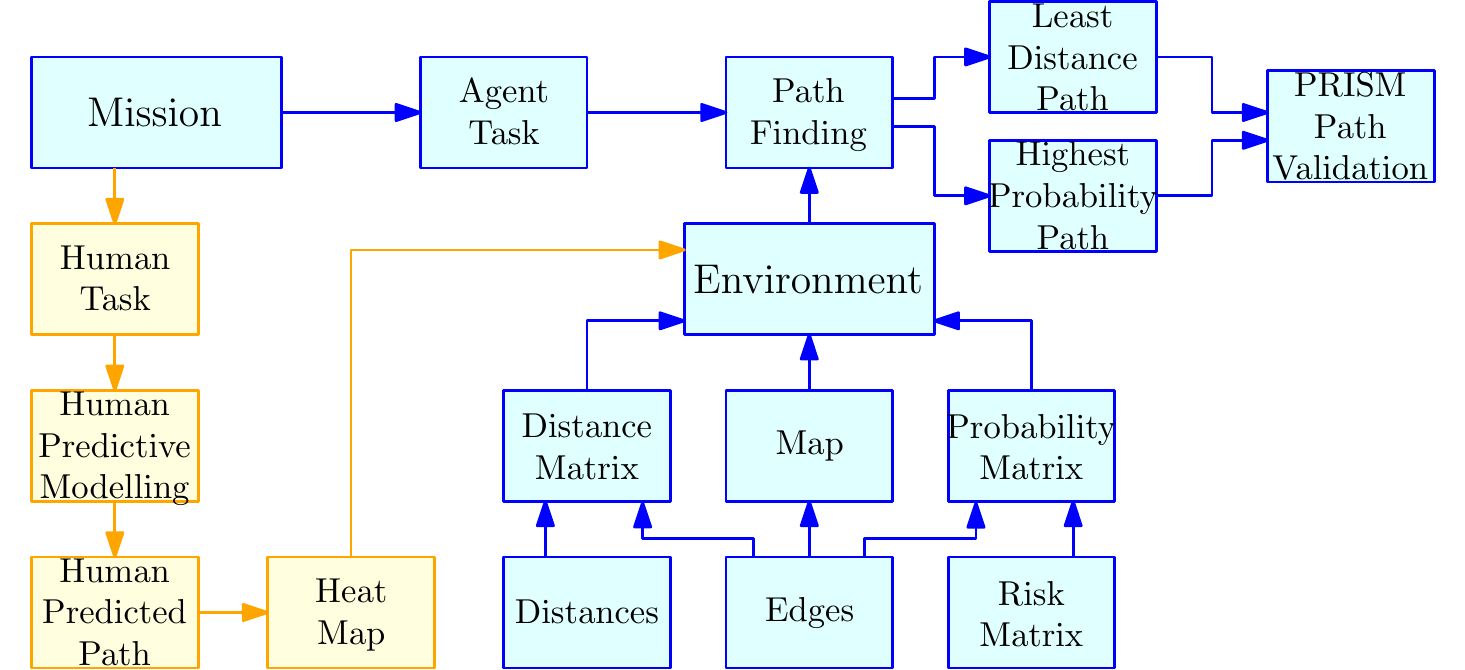}
    \caption{Generalised workflow for modelling of the human and agent.}
    \label{fig:Approach}
    \vspace{-2mm}
\end{figure}

\paragraph{Entity modelling.} \label{subsection: Entity Modelling}
Entity modelling encompasses all modelling associated to the agent and human. 
A generalised workflow of each entity can be seen in Figure~\ref{fig:Approach}, where sections highlighted in yellow are unique solely to the agent's class for performing human predictive modelling and are not active in the human's class. The path finding algorithm for the agent returns two solutions: a path of least distance and a path with the highest probability of success. Whilst the path with the highest probability of success is a good measure of path safety, the path finding algorithm does not systematically determine the probability of success, and therefore an interface enabling the invocation of the probabilistic model checker PRISM~\cite{prism} has been developed. This interface creates a series of actions based on an applied path, generating a unique PRISM model. PRISM is then called at run-time to evaluate the applied path and to systematically determine the success probability of traversing the path. 

\revised{Adapting the system to changes that occur at run-time is a challenging problem because multiple dependencies must be taken into account, as well as trade-offs and uncertainties. Considering all these aspects in planning for adaptation often yields large solution spaces which cannot be adequately explored at run-time. Thus, we employ the path finding algorithm in combination with validation through probabilistic model checking (using the model checker PRISM) for efficiency reasons, given that synthesising paths as part of a Markov decision process (MDP) policy is rather costly.}

\paragraph{Environment modelling.} \label{subsection: Environment Modelling} We model the environment as a graph $\mathcal{G} = (P, E)$ with nodes $P=\{p_1,\ ...,\ p_n\}$ and edges $E$, where each edge $e\in E$ is a tuple $e=(p_i, p_j, D, R)\in P \times P \times \mathbb{R} \times {C_R}$. In this tuple, $D \in \mathbb{R}$ gives the distance between the notes $p_i$ and $p_j$, and $R\in C_R$ is a risk level that can take one of several discrete values. Each risk level $R$ corresponds to a tuple of probabilities $f_R(R)=(P_{s},P_{r},P_{f})$, where $(P_{s})$ is the probability of the agent successfully traversing the edge, $(P_r)$ is the probability of the agent failing to traverse the edge and staying in the same location \revised{temporarily}, and $P_{f}=1-(P_{s}+P_{r})$ is the probability of the robot suffering a catastrophic failure and ending the mission.




\paragraph{Path finding.} \label{subsection: Path Finding}
\begin{wrapfigure}[17]{r}{0.5\textwidth}
\vspace{-21pt}
\begin{minipage}{0.5\textwidth}
    \begin{algorithm}[H]
    \caption{Path finding, validation and selection}
    \label{Algorithm: Path Validation Process}
    \begin{algorithmic}[1]
    \Function{PathValidation}{$\mathcal{G},\mathit{start}, \mathit{final}, \mathcal{G}_h=None$}
        \Statex \hspace{4pt} $\triangleright$ \textsc{Part 1: Path finding}
        \If{$\mathcal{G}_h.\textsc{Exists()}$}
            \State $\mathcal{G} \gets \mathcal{G}_h$ \Comment{Set environment map to $\mathcal{G}_h$}
        \EndIf
        \State $\mathit{path\_dist} \gets \textsc{Dijkstra}(\mathit{start}, \mathit{final}, \mathcal{G})$
        \State $\mathit{path\_prob} \gets \textsc{OptimalPath}(\mathit{start}, \mathit{final}, \mathcal{G})$
        \Statex
        \Statex \hspace{4pt} $\triangleright$ \textsc{Part 2: Path validation}
        \State $(\mathit{p\_dist}, \mathit{p\_prob}) \gets \textsc{Policy}(\mathit{path\_dist}, \mathit{path\_prob})$
        \State $(\mathit{m\_dist}, \mathit{m\_prob}) \gets \textsc{Model}(\mathit{p\_dist}, \mathit{p\_prob})$
        \State $(\mathit{r\_dist}, \mathit{r\_prob}) \gets \textsc{Evaluate}(\mathit{m\_dist}, \mathit{m\_prob})$
        \Statex
        \Statex \hspace{4pt} $\triangleright$ \textsc{Part 3: Path selection}
        \If{$\mathit{r\_dist} > \mathit{r\_prob}$}
            \State $path \gets path\_dist$
        \Else
            \State $path \gets path\_prob$
        \EndIf
        \State \Return $path$
  \EndFunction
  \end{algorithmic}
  \end{algorithm}
 \end{minipage}
  \end{wrapfigure}
Our framework\footnote{\url{https://github.com/hysterr/York_Abstract_Navigation}} performs path finding using two variants of Dijkstra's algorithm, as demonstrated in Algorithm~\ref{Algorithm: Path Validation Process}. First, a conventional Dijkstra's algorithm is used to find the path between $p_i$ and $p_j$ with the shortest distance (line~5). Second, a modified Dijkstra's algorithm is used to find the optimal path against a cost function which maximises the probability of successfully reaching the end state (line~6). 

The function \textsc{PathValidation} requires the default environment map $\mathcal{G}$ as an input argument. A second heated environment map $\mathcal{G}_h$ can be given as an optional argument, which is only used 
when the robot is operating in the presence of a human. 
The location of the human 
and knowledge of their intentions are used to estimate their trajectory through the environment during any stage of the mission. 
Since path finding can be performed for solo missions, the function \textsc{PathValidation} uses the default environment unless a heated environment is specified (lines~2--4). 

Systematic validation of the found paths is performed by creating a probabilistic model \revised{(MDP)} at run-time and evaluating the paths using a fixed-policy. Actions are selected by creating a function which interprets both paths (line 7) and given a specific state, creates a fixed policy which navigates the robot towards the next position along the path. This policy, along with the environment map containing all nodes and edges, is used to create a Markov decision process model of the system (line 8) for both paths. Using the PRISM command-line interface, the PCTL relationship $P_{max}=?\ \textrm{[F\ (end\ \&\ state\ =\ final)]}$, finds the maximum probability of reaching the end state at some point in the future based on the actions selected by the fixed policy (line 9). This returns two values ($r\_dist$ and $p\_dist$) which represent the maximum probability of reaching the end state based on the environment and the fixed policy. These two values are used to select the most appropriate path produced by the dual Dijkstra analysis (lines 10-14). 

\begin{figure}[!t]%
\centering
\subfloat[Conflicted agent and human paths]{\label{fig:Deconflicted Paths}\includegraphics[width=0.42\textwidth]{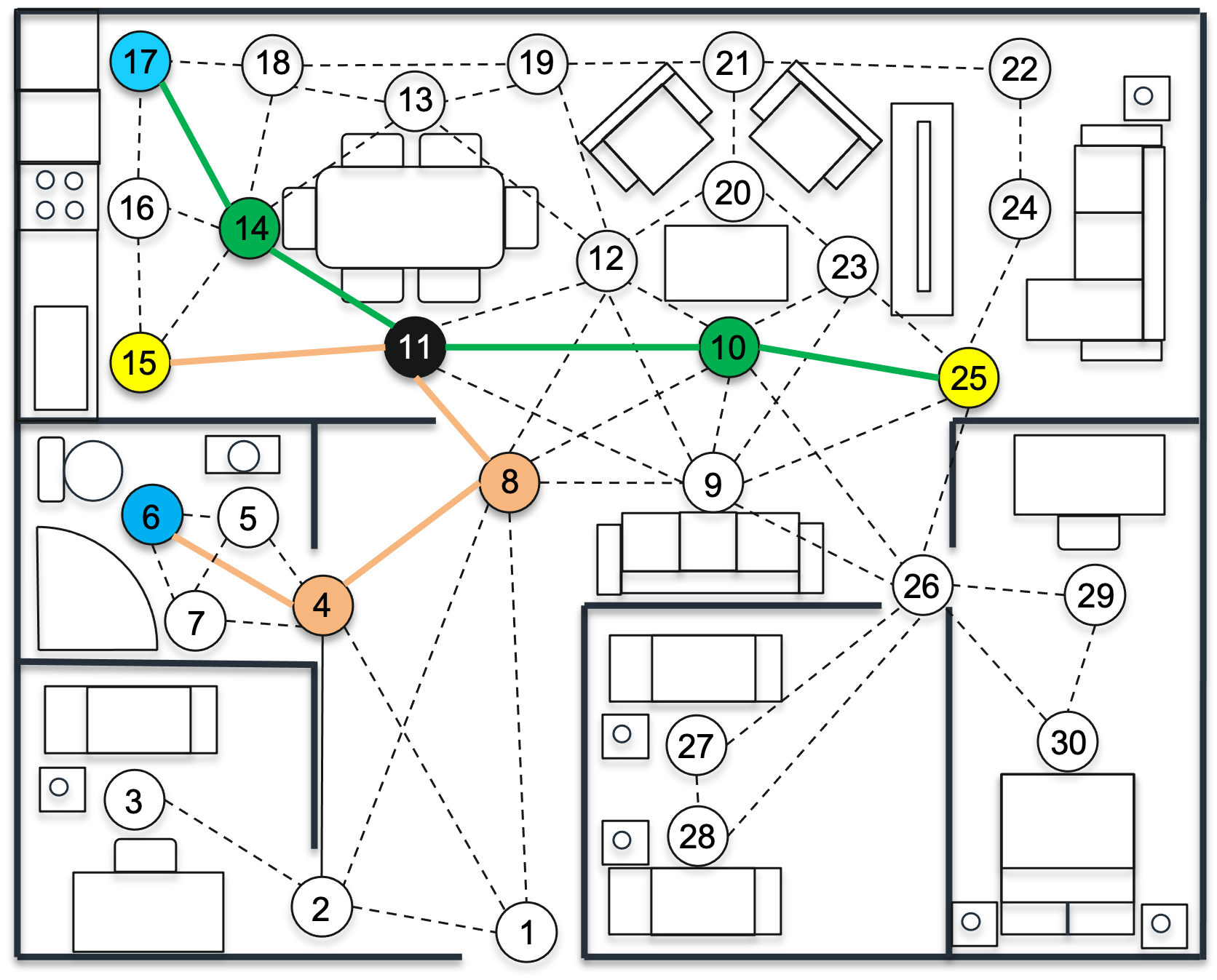}}\qquad
\subfloat[Updated path for the agent]{\label{fig:Updated Path}\includegraphics[width=0.42\textwidth]{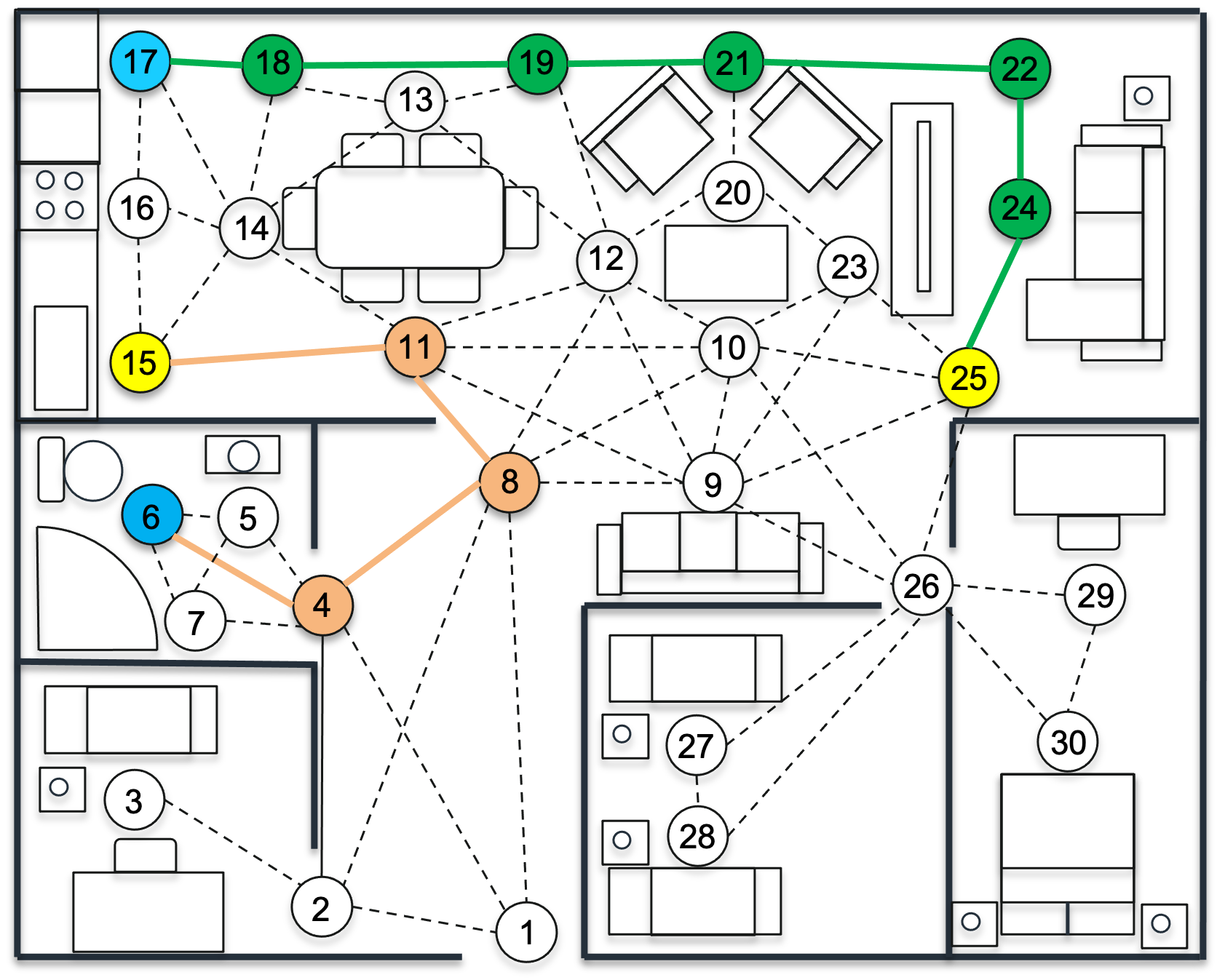}}
\caption{Graphical demonstration of our path-finding framework, where \protect\circledSTART{15} and \protect\circledEND{6} are the start and end nodes for the human, \protect\circledSTART{25} and \protect\circledEND{17} are the start and end nodes for the agent, the node sequence \protect\circledHuman{11} $\rightarrow$ \protect\circledHuman{8} $\rightarrow$ \protect\circledHuman{4} forms the human path, the node sequences \protect\circledAgent{10} $\rightarrow$ \protect\circledAgent{11} $\rightarrow$ \protect\circledAgent{14} and \protect\circledAgent{24} $\rightarrow$ \protect\circledAgent{22} $\rightarrow$ \protect\circledAgent{21} $\rightarrow$ \protect\circledAgent{19} $\rightarrow$ \protect\circledAgent{18} form the initial and updated agent paths, respectively, and \protect\circledConflict{11} is a conflicted node.}
\label{fig:FrameworkDemo}
 \vspace{-2mm}
\end{figure}

Figure~\ref{fig:FrameworkDemo} illustrates our path-finding framework \revised{in an example where the agent navigates through the environment to reach a specified location. The predicted path taken by the human causes a spatial conflict with the highest probability path taken by the agent as both entities at some point in time will be located at node 11. This impacts safety as the agent may be aligned on a trajectory in close proximity to the human or cause distress by blocking their path. However, by using the workflow outlined in }Figure~\ref{fig:Approach}, \revised{the predicted path of the human is interpreted by the agent's representation of the environment in the form of a heat map. This heat map updates the agent's risk matrix, artificially increasing the risk associated with edges of commonality and decreasing the probability of successfully traversing the edge due to the presence of the human. Path finding is then performed again for the agent based on this updated interpretation of the environment.}

Since the predicted path for the human was spatially conflicted with the original highest probability of success path for the agent (Figure~\ref{fig:Deconflicted Paths}), the obtained success probability no longer holds true due to the increased risk from the human presence (97\% to 40\% drop). Therefore, when performing path finding for the agent using the updated environment representation, a different solution is returned corresponding to the highest probability of success. This can be seen in Figure~\ref{fig:Updated Path} where 
an updated path of highest probability guides the agent in avoiding the cluttered central area taken by the least distance path, with approximately 96\% probability of success.

\paragraph{Mission planning.} \label{subsection: Mission Planning}
Missions are defined as a series of tasks, represented by nodes in the environment. A local task is defined as an action or a series of actions performed by the robot accounted for by the inner loop control system embedded on the robot. Such local tasks include identification of objects and manipulation of objects during an object retrieval mission. The order of each action and the complexity of operation for each local task is not considered in the safety case or evaluation phase of this framework, with only navigational complexity of the environment considered in the final solution. 

\paragraph{Human-robot interaction and adaptation.}
Since ensuring the safety of the human is the most important aspect of this research, the robot should adapt its plan at run-time to account for the location and movement of the human. 
%
Besides avoiding traversing through an area where the human is located, the robot can also make requests to the human to perform specific tasks within the mission. This is under the assumption that the human will a) cooperate and begin performing the designated task and b) navigate through the environment following the shortest path, which can be predicted using Dijkstra's algorithm \revised{outlined earlier, but with a modified cost function which minimises the distance travelled as opposed to decreasing risk. The path predicted for the human is applied to path planning for the robot by adding a linear risk scaling factor to edges that could cause a positional conflict between the human and robot.} 

Additionally, to capture the uncertainty that is typically associated with the human~\cite{9196226,10.1145/3487921}, a variable is assigned to the human representing the likelihood of unpredictable behaviour. To account for this, a secondary evaluation of the edges in the graph that connect to the human's current position is performed to identify where the human could move in the next step in case of diverging from the predicted path.

\section{Evaluation}
The main focus of the evaluation is to assess the performance of the path planning and mission planning framework which aims in maximising the robot reaching the end state, while increasing the uncertainty value of the human during a single phase mission. The uncertainty of the human is a measure of how predictable the human’s movements are, where uncertainty $\in[0,1]$. As uncertainty tends towards zero, the human movements become predictable 
and as it tends towards one, they become unpredictable.

A single-phase mission is comprised by multiple un-ordered tasks with only a single ordered task, the end state. This means that 
it is up to the mission planning framework to determine the optimal mission plan. To test the uncertainty value of the human, the tasks within the mission remained constant throughout the duration of the test, with the mission tasks designed to replicate a surveillance mission whereby the robot navigates the entirety of the multi-room environment introduced in Section~2.

The tasks for this mission are $3,6,9,12,16,24,29$ with $22$ being the end location for the robot. The human is also present in the environment, with their initial location being randomly computed at the start of the simulation in the same way the robot is initialised. As previously mentioned, the robot can request human movement 
if the robot’s path is blocked, or if there is too much uncertainty for the robot to safely move. This redirect request is performed by requesting the human move to a predefined \textit{safe location}. These safe locations are locations within the
environment where the human can rest without blocking the path of the robot. In our case study, these safe locations are $13,14,20,24$.

The uncertainty was evaluated in the range of $[0,1]$ in increments of $0.1$, with each increment evaluated over $25,000$ episodes. Each assessed episode was given a pass or fail using the following criteria:

\begin{enumerate}
    \item If the agent reaches the end of the mission, the episode was a success.
    \item If the agent engages a hold state for $10$ consecutive steps, the mission ends with a failure.
    \item If the agent fails due to an uncertain probabilistic action, the mission ends with a failure.
\end{enumerate}

\begin{table}[!t]
\sffamily
\centering
\caption{Success, failure and redirects based on uncertainty.}
\label{table:results}
\begin{footnotesize}
\begin{tabular}{ccccccc}
\toprule
\multirow{2}{*}{Uncertainty} & \multirow{2}{*}{\shortstack[c]{Success (\%)}} & \multirow{2}{*}{\shortstack[c]{Success \\Rate}} & \multirow{2}{*}{\shortstack[c]{Fail \\Rate}} & \multirow{2}{*}{\shortstack[c]{Total \\Redirects}} & \multirow{2}{*}{\shortstack[c]{Redirect (\%)}} & \multirow{2}{*}{\shortstack[c]{Max Redirect \\Requests}} \\
             &              &                &             &            &              &           \\ \midrule
       0     &  99.82\%     &     24955      &     45      &    11369   &   45.48\%    &     5      \\
       0.1   &  99.77\%     &     24942      &     58      &    12799   &   51.20\%    &     5     \\
       0.2   &  99.31\%     &     24872      &     173     &    13193   &   52.68\%    &     5      \\
       0.3   &  98.63\%     &     24657      &     343     &    13391   &   53.56\%    &     6      \\
       0.4   &  97.58\%     &     24396      &     604     &    13113   &   52.45\%    &     6      \\
       0.5   &  95.92\%     &     23981      &     1019    &    12672   &   50.69\%    &     6      \\
       0.6   &  94.13\%     &     23533      &     1467    &    12080   &   48.31\%    &     5      \\
       0.7   &  91.75\%     &     22937      &     2063    &    11138   &   44.55\%    &     4      \\
       0.8   &  89.17\%     &     22293      &     2707    &    9863    &   39.45\%    &     3      \\
       0.9   &  88.44\%     &     22109      &     2891    &    8338    &   33.35\%    &     2      \\
        1    &  88.54\%     &     22104      &     2860    &    6077    &   24.34\%    &     1      \\ \bottomrule
\end{tabular}
\end{footnotesize}
 \vspace{-2mm}
\end{table}

Based on the resulting data in Table~\ref{table:results}, as the uncertainty of the human increases, the mission completion success rate gradually declines. 
With $0$ uncertainty, $\approx$45\% of episodes required at least one redirect. This was expected since the human was randomly initialised in the environment and with nine nodes acting as task nodes, there was $\approx$30\% chance that the human would be initially located at one of these nodes and would require a redirect. The other $15\%$ is attributed to scenarios where the human and robot were both initialised at the same node or the human was obstructing the path of the robot.

When analysing the number of redirect requests during successful missions, if a mission had one or two redirects, the mission was always successfully completed. This implies that the redirects were successfully triggered and the human complied by taking the predicted paths. However, as the number of redirect requests per episode exceeded two, the number of failures started to increase. Upon analysing the specific missions causing these multiple redirects and failure states, it was evident that all had a common factor, the human located centrally in the environment. When the robot operates in this region, it creates a larger uncertainty, as the human could move in multiple directions. This caused the robot to engage its hold state and begin requesting human redirects as the associated risk fell below the set $0.9$ threshold.

As the uncertainty value increased, the human was more likely to move away from the central region. 
For values $0.3$ and $0.4$, the likelihood of a mission succeeding with a high number of requests increased substantially (e.g., episodes with five requests for $0.3$ uncertainty had $83\%$ success rate, compared to the $3\%$ success rate for $0$ uncertainty). When uncertainty exceeded $0.4$, the human became more unpredictable, reducing the likelihood of a human-robot encounter along a path \revised{and requesting a redirect}. Despite the erratic behaviour of the human, missions continued to reduce the maximum number of requests made as the human would directly encounter the robot less frequently, with only $24\%$ of episodes having at \revised{most} one redirect as the uncertainty approached $1$. Despite a lower number of requests overall and per episode, the success rate of the robot continued to decline proportionally to the uncertainty, as the human behaviour led the threshold value to drop below $0.9$ which was required for the robot to move to the next node. 

It is evident that there is a fine balance between what is deemed acceptable regarding the human uncertainty. A high success rate (over $98\%$) is achieved when the human moves only under request. However, when operating inside a heavily connected region, 
the movement of the robot can become stagnated due to a high threshold value ($0.9$) for selecting a movement. As the uncertainty of the human started to increase, the likelihood that the human would remain within this heavily connected region reduced, increasing the probability that an episode which had numerous requests would still remain successful. Despite this, the number of successfully completed missions continued to decline as the human’s unpredictable movement lowered the probability of successfully traversing an edge due to increased uncertainty. It is expected that by lowering the threshold value for the robot would allow the robot to select more risky and uncertain actions, but would also increase the chance of achieving a failure state.


\section{Related Work}

In recent work~\cite{gleirscher2022verified,cobotS}, we devised a method for synthesising Pareto-optimal safety controllers to ensure the safety of human operators in collaborative tasks with robots in a manufacturing setting. However, this research was established under the assumption that the operator is a trained user and uncertainty arising due to unpredictable human behaviour is not a critical factor, contrary to this work.   
%

A Markov decision process model was created in~\cite{Lacerda_2019,DBLP:conf/aips/Lacerda0H17} representing a social robot operating in uncertain environment, with optimal policy generation guaranteeing formal performance through probabilistic verification techniques. This work aims in maximising the success probability of tasks while minimising the expected time and cost. 
Uncertainties within the environment are modelled probabilistically in~\cite{Lu2020,Tihanyi2021}, with paths generated in a grid-world environment, maximising safety while producing computationally tractable solutions. This approach focuses on safe operation in hazardous environments.  

Social human-robot interaction is introduced in~\cite{Hoffman2004}, enabling intuitive and natural communication with a human and cooperating as partners in a collaborative mission. The robot does not perform mission management in the traditional sense, but maintains a repertoire of tasks represented as a hierarchical structure of actions and sub-tasks, arranged into an action tuple data structure. 
Human-robot collaboration is presented in~\cite{Karami2010} towards completing a common mission. Without knowledge of each others plan, the robot builds a belief of the human intentions 
by observing their actions. Robot decisions are supported by the use of partially observable Markov decision processes. However, the demonstrations are in extremely benign situations with the human and robot located in a small and clutter-free environment. 

\revised{The work presented in} \cite{Camara20} \revised{proposes a planning approach that assesses the impact of mutual dependencies between software architecture and task planning on the satisfaction of mission goals. Similarly to our approach, the authors employ Dijkstra's algorithm to compute and rank paths in ascending distance to reduce the computational cost compared to considering the full solution space in a single model checking run. This work differs from ours as it focuses on architecture reconfiguration and it does not take into account potential interactions with humans. Self-adaptation in autonomous robots is also the topic in} \cite{8787014}. \revised{The difference with the work in }\cite{8787014} \revised{and with our approach is that it uses machine learning to find Pareto-optimal configurations without needing to explore every configuration and restricts the search space to such configurations to make planning tractable, but again, without considering any potential interactions with humans.}

\section{Conclusion}

We presented work in progress to develop an adaptive path planning framework for robotic mission execution with active human presence. We illustrated its use in a simulated assistive-care case study in which a mobile robot monitors an end user with mild physical or cognitive impairments. In future work, we plan to introduce multi-phase missions (i.e., missions with tasks changing over time), support re-planning (i.e., re-ordering of remaining tasks to eliminate human interference with the robot's progress) instead of requesting redirects, and finally, evaluate the framework in additional environments.

\bibliographystyle{eptcs}
\bibliography{references}

\begin{thebibliography}{10}
\providecommand{\bibitemdeclare}[2]{}
\providecommand{\surnamestart}{}
\providecommand{\surnameend}{}
\providecommand{\urlprefix}{Available at }
\providecommand{\url}[1]{\texttt{#1}}
\providecommand{\href}[2]{\texttt{#2}}
\providecommand{\urlalt}[2]{\href{#1}{#2}}
\providecommand{\doi}[1]{doi:\urlalt{https://doi.org/#1}{#1}}
\providecommand{\eprint}[1]{arXiv:\urlalt{https://arxiv.org/abs/#1}{#1}}
\providecommand{\bibinfo}[2]{#2}

\bibitemdeclare{article}{Bedaf2017}
\bibitem{Bedaf2017}
\bibinfo{author}{Sandra \surnamestart Bedaf\surnameend},
  \bibinfo{author}{Patrizia \surnamestart Marti\surnameend},
  \bibinfo{author}{Farshid \surnamestart Amirabdollahian\surnameend} \&
  \bibinfo{author}{Luc \surnamestart de~Witte\surnameend}
  (\bibinfo{year}{2017}): \emph{\bibinfo{title}{A multi-perspective evaluation
  of a service robot for seniors: the voice of different stakeholders}}.
\newblock {\slshape \bibinfo{journal}{Disability and Rehabilitation: Assistive
  Technology}} \bibinfo{volume}{13}(\bibinfo{number}{6}), pp.
  \bibinfo{pages}{592--599}, \doi{10.1080/17483107.2017.1358300}.

\bibitemdeclare{article}{Beuscher2017}
\bibitem{Beuscher2017}
\bibinfo{author}{Linda~M. \surnamestart Beuscher\surnameend},
  \bibinfo{author}{Jing \surnamestart Fan\surnameend},
  \bibinfo{author}{Nilanjan \surnamestart Sarkar\surnameend},
  \bibinfo{author}{Mary~S. \surnamestart Dietrich\surnameend},
  \bibinfo{author}{Paul~A. \surnamestart Newhouse\surnameend},
  \bibinfo{author}{Karen~F. \surnamestart Miller\surnameend} \&
  \bibinfo{author}{Lorraine~C. \surnamestart Mion\surnameend}
  (\bibinfo{year}{2017}): \emph{\bibinfo{title}{{Socially assistive robots:
  Measuring older adults' perceptions}}}.
\newblock {\slshape \bibinfo{journal}{Journal of Gerontological Nursing}}
  \bibinfo{volume}{43}(\bibinfo{number}{12}), pp. \bibinfo{pages}{35--43},
  \doi{10.3928/00989134-20170707-04}.

\bibitemdeclare{inproceedings}{calinescu2013emerging}
\bibitem{calinescu2013emerging}
\bibinfo{author}{Radu \surnamestart Calinescu\surnameend}
  (\bibinfo{year}{2013}): \emph{\bibinfo{title}{Emerging techniques for the
  engineering of self-adaptive high-integrity software}}.
\newblock In: {\slshape \bibinfo{booktitle}{Assurances for Self-Adaptive
  Systems}}, \bibinfo{publisher}{Springer}, pp. \bibinfo{pages}{297--310},
  \doi{10.1007/978-3-642-36249-1\_11}.

\bibitemdeclare{inproceedings}{calinescu2019socio}
\bibitem{calinescu2019socio}
\bibinfo{author}{Radu \surnamestart Calinescu\surnameend},
  \bibinfo{author}{Javier \surnamestart C{\'a}mara\surnameend} \&
  \bibinfo{author}{Colin \surnamestart Paterson\surnameend}
  (\bibinfo{year}{2019}): \emph{\bibinfo{title}{Socio-cyber-physical systems:
  Models, opportunities, open challenges}}.
\newblock In: {\slshape \bibinfo{booktitle}{Software Engineering for Smart
  Cyber-Physical Systems}}, pp. \bibinfo{pages}{2--6},
  \doi{10.1109/SEsCPS.2019.00008}.

\bibitemdeclare{inproceedings}{9196226}
\bibitem{9196226}
\bibinfo{author}{Radu \surnamestart Calinescu\surnameend},
  \bibinfo{author}{Raffaela \surnamestart Mirandola\surnameend},
  \bibinfo{author}{Diego \surnamestart Perez-Palacin\surnameend} \&
  \bibinfo{author}{Danny \surnamestart Weyns\surnameend}
  (\bibinfo{year}{2020}): \emph{\bibinfo{title}{Understanding Uncertainty in
  Self-adaptive Systems}}.
\newblock In: {\slshape \bibinfo{booktitle}{2020 IEEE International Conference
  on Autonomic Computing and Self-Organizing Systems (ACSOS)}}, pp.
  \bibinfo{pages}{242--251}, \doi{10.1109/ACSOS49614.2020.00047}.

\bibitemdeclare{inproceedings}{Camara20}
\bibitem{Camara20}
\bibinfo{author}{Javier \surnamestart C\'{a}mara\surnameend},
  \bibinfo{author}{Bradley \surnamestart Schmerl\surnameend} \&
  \bibinfo{author}{David \surnamestart Garlan\surnameend}
  (\bibinfo{year}{2020}): \emph{\bibinfo{title}{Software Architecture and Task
  Plan Co-Adaptation for Mobile Service Robots}}.
\newblock In: {\slshape \bibinfo{booktitle}{Proceedings of the IEEE/ACM 15th
  International Symposium on Software Engineering for Adaptive and
  Self-Managing Systems}}, \bibinfo{publisher}{Association for Computing
  Machinery}, p. \bibinfo{pages}{125–136}, \doi{10.1145/3387939.3391591}.

\bibitemdeclare{inproceedings}{FeilSeifer}
\bibitem{FeilSeifer}
\bibinfo{author}{D.~\surnamestart Feil-Seifer\surnameend} \&
  \bibinfo{author}{M.J. \surnamestart Mataric\surnameend}
  (\bibinfo{year}{2005}): \emph{\bibinfo{title}{Defining socially Assistive
  Robotics}}.
\newblock In: {\slshape \bibinfo{booktitle}{9th International Conference on
  Rehabilitation Robotics (ICORR)}}, \bibinfo{publisher}{{IEEE}}, pp.
  \bibinfo{pages}{465--468}, \doi{10.1109/icorr.2005.1501143}.

\bibitemdeclare{article}{gleirscher2022verified}
\bibitem{gleirscher2022verified}
\bibinfo{author}{Mario \surnamestart Gleirscher\surnameend},
  \bibinfo{author}{Radu \surnamestart Calinescu\surnameend},
  \bibinfo{author}{James \surnamestart Douthwaite\surnameend},
  \bibinfo{author}{Benjamin \surnamestart Lesage\surnameend},
  \bibinfo{author}{Colin \surnamestart Paterson\surnameend},
  \bibinfo{author}{Jonathan \surnamestart Aitken\surnameend},
  \bibinfo{author}{Rob \surnamestart Alexander\surnameend} \&
  \bibinfo{author}{James \surnamestart Law\surnameend} (\bibinfo{year}{2022}):
  \emph{\bibinfo{title}{Verified synthesis of optimal safety controllers for
  human-robot collaboration}}.
\newblock {\slshape \bibinfo{journal}{Science of Computer Programming}}
  \bibinfo{volume}{218}, p. \bibinfo{pages}{102809},
  \doi{10.1016/j.scico.2022.102809}.

\bibitemdeclare{inproceedings}{Gleirscher2022}
\bibitem{Gleirscher2022}
\bibinfo{author}{Mario \surnamestart Gleirscher\surnameend},
  \bibinfo{author}{Nikita \surnamestart Johnson\surnameend},
  \bibinfo{author}{Panayiotis \surnamestart Karachristou\surnameend},
  \bibinfo{author}{Radu \surnamestart Calinescu\surnameend},
  \bibinfo{author}{James \surnamestart Law\surnameend} \& \bibinfo{author}{John
  \surnamestart Clark\surnameend} (\bibinfo{year}{2022}):
  \emph{\bibinfo{title}{Challenges in the Safety-Security Co-Assurance of
  Collaborative Industrial Robots}}.
\newblock In: {\slshape \bibinfo{booktitle}{The 21st Century Industrial Robot:
  When Tools Become Collaborators}}, \bibinfo{publisher}{Springer}, pp.
  \bibinfo{pages}{191--214}, \doi{10.1007/978-3-030-78513-0\_11}.

\bibitemdeclare{article}{Hebesberger2017}
\bibitem{Hebesberger2017}
\bibinfo{author}{Denise \surnamestart Hebesberger\surnameend},
  \bibinfo{author}{Tobias \surnamestart Koertner\surnameend},
  \bibinfo{author}{Christoph \surnamestart Gisinger\surnameend} \&
  \bibinfo{author}{J\"{u}rgen \surnamestart Pripfl\surnameend}
  (\bibinfo{year}{2017}): \emph{\bibinfo{title}{A Long-Term Autonomous Robot at
  a Care Hospital: A Mixed Methods Study on Social Acceptance and Experiences
  of Staff and Older Adults}}.
\newblock {\slshape \bibinfo{journal}{International Journal of Social
  Robotics}} \bibinfo{volume}{9}(\bibinfo{number}{3}), pp.
  \bibinfo{pages}{417--429}, \doi{10.1007/s12369-016-0391-6}.

\bibitemdeclare{article}{10.1145/3487921}
\bibitem{10.1145/3487921}
\bibinfo{author}{Sara~M. \surnamestart Hezavehi\surnameend},
  \bibinfo{author}{Danny \surnamestart Weyns\surnameend},
  \bibinfo{author}{Paris \surnamestart Avgeriou\surnameend},
  \bibinfo{author}{Radu \surnamestart Calinescu\surnameend},
  \bibinfo{author}{Raffaela \surnamestart Mirandola\surnameend} \&
  \bibinfo{author}{Diego \surnamestart Perez-Palacin\surnameend}
  (\bibinfo{year}{2021}): \emph{\bibinfo{title}{Uncertainty in Self-Adaptive
  Systems: A Research Community Perspective}}.
\newblock {\slshape \bibinfo{journal}{ACM Trans. Auton. Adapt. Syst.}}
  \bibinfo{volume}{15}(\bibinfo{number}{4}), \doi{10.1145/3487921}.

\bibitemdeclare{article}{Hoffman2004}
\bibitem{Hoffman2004}
\bibinfo{author}{Guy \surnamestart Hoffman\surnameend} \&
  \bibinfo{author}{Cynthia \surnamestart Breazeal\surnameend}
  (\bibinfo{year}{2004}): \emph{\bibinfo{title}{{Robots that work in
  collaboration with people}}}.
\newblock {\slshape \bibinfo{journal}{AAAI Fall Symposium - Technical Report}}
  \bibinfo{volume}{FS-04-05}, pp. \bibinfo{pages}{25--30}.

\bibitemdeclare{inproceedings}{8787014}
\bibitem{8787014}
\bibinfo{author}{Pooyan \surnamestart Jamshidi\surnameend},
  \bibinfo{author}{Javier \surnamestart Cámara\surnameend},
  \bibinfo{author}{Bradley \surnamestart Schmerl\surnameend},
  \bibinfo{author}{Christian \surnamestart Käestner\surnameend} \&
  \bibinfo{author}{David \surnamestart Garlan\surnameend}
  (\bibinfo{year}{2019}): \emph{\bibinfo{title}{Machine Learning Meets
  Quantitative Planning: Enabling Self-Adaptation in Autonomous Robots}}.
\newblock In: {\slshape \bibinfo{booktitle}{IEEE/ACM 14th International
  Symposium on Software Engineering for Adaptive and Self-Managing Systems
  (SEAMS)}}, pp. \bibinfo{pages}{39--50}, \doi{10.1109/SEAMS.2019.00015}.

\bibitemdeclare{inproceedings}{Karami2010}
\bibitem{Karami2010}
\bibinfo{author}{Abir-Beatrice \surnamestart Karami\surnameend},
  \bibinfo{author}{Laurent \surnamestart Jeanpierre\surnameend} \&
  \bibinfo{author}{Abdel-Illah \surnamestart Mouaddib\surnameend}
  (\bibinfo{year}{2010}): \emph{\bibinfo{title}{Human-robot collaboration for a
  shared mission}}.
\newblock In: {\slshape \bibinfo{booktitle}{5th {ACM}/{IEEE} International
  Conference on Human-Robot Interaction}}, pp. \bibinfo{pages}{155--156},
  \doi{10.1109/HRI.2010.5453219}.

\bibitemdeclare{inproceedings}{prism}
\bibitem{prism}
\bibinfo{author}{Marta \surnamestart Kwiatkowska\surnameend},
  \bibinfo{author}{Gethin \surnamestart Norman\surnameend} \&
  \bibinfo{author}{David \surnamestart Parker\surnameend}
  (\bibinfo{year}{2011}): \emph{\bibinfo{title}{PRISM 4.0: Verification of
  Probabilistic Real-Time Systems}}.
\newblock In: {\slshape \bibinfo{booktitle}{Computer Aided Verification}}, pp.
  \bibinfo{pages}{585--591}, \doi{10.1007/978-3-642-22110-1\_47}.

\bibitemdeclare{article}{Lacerda_2019}
\bibitem{Lacerda_2019}
\bibinfo{author}{Bruno \surnamestart Lacerda\surnameend},
  \bibinfo{author}{Fatma \surnamestart Faruq\surnameend},
  \bibinfo{author}{David \surnamestart Parker\surnameend} \&
  \bibinfo{author}{Nick \surnamestart Hawes\surnameend} (\bibinfo{year}{2019}):
  \emph{\bibinfo{title}{Probabilistic planning with formal performance
  guarantees for mobile service robots}}.
\newblock {\slshape \bibinfo{journal}{International Journal of Robotics
  Research}} \bibinfo{volume}{38}(\bibinfo{number}{9}), pp.
  \bibinfo{pages}{1098--1123}, \doi{10.1177/0278364919856695}.

\bibitemdeclare{inproceedings}{DBLP:conf/aips/Lacerda0H17}
\bibitem{DBLP:conf/aips/Lacerda0H17}
\bibinfo{author}{Bruno \surnamestart Lacerda\surnameend},
  \bibinfo{author}{David \surnamestart Parker\surnameend} \&
  \bibinfo{author}{Nick \surnamestart Hawes\surnameend} (\bibinfo{year}{2017}):
  \emph{\bibinfo{title}{Multi-Objective Policy Generation for Mobile Robots
  under Probabilistic Time-Bounded Guarantees}}.
\newblock In: {\slshape \bibinfo{booktitle}{27th International Conference on
  Automated Planning and Scheduling}}, pp. \bibinfo{pages}{504--512}.

\bibitemdeclare{inproceedings}{Lu2020}
\bibitem{Lu2020}
\bibinfo{author}{Yimeng \surnamestart Lu\surnameend} \& \bibinfo{author}{Maryam
  \surnamestart Kamgarpour\surnameend} (\bibinfo{year}{2020}):
  \emph{\bibinfo{title}{Safe Mission Planning under Dynamical Uncertainties}}.
\newblock In: {\slshape \bibinfo{booktitle}{IEEE International Conference on
  Robotics and Automation (ICRA)}}, pp. \bibinfo{pages}{2209--2215},
  \doi{10.48550/arXiv.2003.02913}.

\bibitemdeclare{inproceedings}{cobotS}
\bibitem{cobotS}
\bibinfo{author}{Ioannis \surnamestart Stefanakos\surnameend},
  \bibinfo{author}{Radu \surnamestart Calinescu\surnameend},
  \bibinfo{author}{James \surnamestart Douthwaite\surnameend},
  \bibinfo{author}{Jonathan \surnamestart Aitken\surnameend} \&
  \bibinfo{author}{James \surnamestart Law\surnameend} (\bibinfo{year}{2022}):
  \emph{\bibinfo{title}{{Safety Controller Synthesis for a Mobile Manufacturing
  Cobot}}}.
\newblock In: {\slshape \bibinfo{booktitle}{International Conference on
  Software Engineering and Formal Methods}}.
\newblock \bibinfo{note}{{(forthcoming)}}.

\bibitemdeclare{misc}{Tihanyi2021}
\bibitem{Tihanyi2021}
\bibinfo{author}{Daniel \surnamestart Tihanyi\surnameend},
  \bibinfo{author}{Yimeng \surnamestart Lu\surnameend}, \bibinfo{author}{Orcun
  \surnamestart Karaca\surnameend} \& \bibinfo{author}{Maryam \surnamestart
  Kamgarpour\surnameend} (\bibinfo{year}{2021}):
  \emph{\bibinfo{title}{Multi-robot task allocation for safe planning against
  stochastic hazard dynamics}}, \doi{10.48550/ARXIV.2103.01840}.

\end{thebibliography}
\end{document}